\documentclass[11pt]{elsarticle}
\usepackage{placeins}
\usepackage{lineno,hyperref,color}
\usepackage[dvipsnames]{xcolor}
\usepackage{mwe}
\usepackage{footnote}
\usepackage{comment}
\usepackage{subcaption}

\modulolinenumbers[5]
\definecolor{LightCyan}{rgb}{0.5,0.3,0.1}
\definecolor{Gray}{rgb}{0.7,0.8,1}

\usepackage{array}
\newcolumntype{M}[1]{>{\centering\arraybackslash}m{#1}}
\journal{Pattern Recognition}

\usepackage{xcolor}
\usepackage{breqn}
\usepackage{amsmath,amssymb}
\usepackage{ulem}
\usepackage{float}
\usepackage[most]{tcolorbox}
\usepackage{multirow}

\usepackage{pifont}

\begin{document}

\begin{frontmatter}



\title{\textit{ST-KeyS}: Self-Supervised Transformer for Keyword Spotting in Historical Handwritten Documents}

\author[mymainaddress3,mymainaddress1]{Sana Khamekhem Jemni}
\author[mymainaddress3,mymainaddress4]{Sourour Ammar\corref{1}}
\author[mymainaddress2]{Mohamed Ali Souibgui}
\author[mymainaddress3,mymainaddress4]{Yousri Kessentini}
\author[mymainaddress5]{Abbas Cheddad}

\address[mymainaddress3]{Digital Research Center of Sfax, B.P. 275, Sakiet Ezzit, 3021 Sfax, Tunisia}
\address[mymainaddress4]{SM@RTS : Laboratory of Signals, systeMs, aRtificial Intelligence and neTworkS, Sfax, Tunisia}
\address[mymainaddress1]{MIR@CL: Multimedia, InfoRmation systems and Advanced Computing Laboratory}
\address[mymainaddress2]{Computer Vision Center, Computer Science Department, Universitat Autònoma de Barcelona, Spain}
\address[mymainaddress5]{Department of Computer Science, Blekinge Institute of Technology, Karlskrona, Sweden}
 \cortext[1]{Corresponding author}

\begin{abstract}
Keyword spotting (KWS) in historical documents is an important tool  for the initial exploration of digitized collections. Nowadays, the most efficient KWS methods are relying on machine learning techniques that require a large amount of annotated training data. However, in the case of historical manuscripts, there is a lack of annotated corpus for training. To handle the data scarcity issue, we investigate the merits of the self-supervised learning to extract useful representations of the input data without relying on human annotations and then using these representations in the downstream task.
We propose ST-KeyS, a masked auto-encoder model based on vision transformers where the  pretraining stage is based on the mask-and-predict paradigm, without the need of labeled data. In the fine-tuning stage, the pre-trained encoder is integrated into a siamese neural network model that is fine-tuned to improve feature embedding from the input images. We further improve the image representation using pyramidal histogram of characters (PHOC) embedding to create and exploit an intermediate representation of images based on text attributes.
In an exhaustive experimental evaluation on three widely used benchmark datasets (Botany, Alvermann Konzilsprotokolle and George Washington), the proposed approach outperforms state-of-the-art methods trained on the same datasets.


\end{abstract}


\end{frontmatter}

\section{Introduction}

Many digitized historical handwritten documents are hard to access due to the lack of suitable indexing and retrieval tools. One solution is to recognize the image by transforming it to text with an automatic tool, or manually.  However, fully manual transcription is time-consuming and expensive. Also,  current  optical character recognition (OCR) systems are mostly efficient for modern printed documents \cite{Fernandez2007,Frinken2012}. But, these systems suffer difficulties in the case of a historical manuscript that usually contains many types of degradation due to bad paper quality, writing style variations, ink flow, shadows, non-uniform lighting, stains, etc.
 
Another solution to handle the non-indexed documents is the word-matching process, which is based on a low-level matching known as $word$ $spotting$ \cite{giotis2017survey}. It is closely associated with content-based image retrieval, since it searches for a word in a set of non-indexed documents using the query image content as the single source of information. As a result, the system will return a ranked list of document word images to the user according to their similarity to the desired searched word image. Ultimately, word spotting can be defined as the process of identifying positions on a document image that have a high potential to correspond to an instance of a query word image, without recognizing it explicitly.

In the literature, two distinct strategies for word spotting appeared depending on the search space, which could be either a set of isolated/segmented word images (segmentation-based approach) or a full document image (segmentation-free approach). In this work, we assume that document images have been segmented into isolated word images serving thereafter for matching the query word image, following previous studies such as ~\cite{Almazan2014,Krishnan2016,Retsinas2019,riba2021} 
Localizing and retrieving word images has been an active field  of research, with the goal of finding the more effective word representations that  give the most meaningful distances between word images.
Early approaches were  using the classic learning-free techniques that rely on expert designed feature embedding (handcrafted features) ~\cite{Retsinas2019,Vats2019}. These methods tend to be immediately applied since they do not rely on a learning stage.

Later, and following the success of other computer vision tasks, machine learning-based techniques (in particular convolutional neural networks (CNNs)) are now dominating the field of word retrieval ~\cite{Sudholt2018,Wilkinson2017,Serdouk2019,Toselli2019,Daraee2021, de2022few}. But, despite the significant improvement in the performance of these models over classical handcrafted techniques, they have their associated shortcomings. First, CNNs operate on regular grids and use the same convolutional filter to extract features from handwritten word images, making this technique sensitive to rotation. 
Second, CNNs fail to capture relevant features for long-range dependencies, as they are more suited to extract low-level spatial information from images. 

With the recent success of  transformers in natural language processing (NLP) \cite{vaswani2017attention, devlin2018bert}, their application in computer vision tasks (such as object detection \cite{carion2020end}, image recognition \cite{dosovitskiy2021an}, question answering \cite{biten2021latr}, image restoration \cite{Souibgui2022ICPR}, handwritten text recognition (HTR) \cite{Minghao2021,souibgui2022text}, named entity recognition \cite{rouhou2021transformer}, etc.) has also lately been getting more attention. The self-attention mechanism that has been proposed in \cite{vaswani2017attention} allows capturing contextual feature interactions information.  This use of local knowledge combined with the information of the global long-range spatial arrangement is beneficial for an efficient keyword spotting (KWS) model.

However, the main issue of transformers is that they are data-hungry. Nonetheless, huge annotated datasets are hard to obtain and labeling large amounts of data is work intensive and can be very expensive, while unlabeled data is much more abundant.  Thus, the need to build models that benefit from the unlabeled data in addition to the annotated data, or even omitting the use of any annotated data becomes more appropriate in this situation. In this regard, self-supervised techniques based on ViT (Vision Transformer) models have recently shown significant results.
 
Motivated by this success  in computer vision, such as image classification and object detection \cite{MAEVit, bao2021beit}, image retrieval \cite{Jang2021} and speech recognition \cite{Sadhu2021} tasks, we propose in this paper an end-to-end keyword spotting approach in handwritten documents which is based on a self-supervised technique and makes use of masked autoencoders with the self-attention mechanism. To the best of our knowledge, this is the first work that proposes a self supervised learning paradigm of transformer based model in the context of word spotting in historical document images.
Our framework is built in two stages. The first stage is the pretraining which is designed for learning useful representations from the unlabeled data, using a masked encoder-decoder architecture, in a self-supervised way. The second stage is the fine-tuning which is devoted to efficiently extracting relevant features from the labeled word image. In the fine-tuning stage, the pretrained encoder is integrated to a siamese neural network (SNN) and is fine-tuned using a few labeled data to improve feature embeddings from the input text images. Then, the resulting model is used as a core component of our word spotting framework by including contextual information extracted from the word text. To achieve this, we align the visual representations extracted by our encoder with the  pyramidal histogram of characters (PHOC) representation.  To demonstrate the effectiveness of the proposed method, we conduct several experiments using various training conditions and several public image databases.

The overall contributions of this paper can be summarized as follows:
\begin{itemize}
    
    \item 
    To the best of our knowledge, we present the first self-supervised approach for the goal of keyword spotting in handwritten text images, composed of pretraining and fine-tuning stages. The approach is  based on vision transformers in an encoder-decoder fashion, without any dependency on CNNs.
    
    \item An effective pretext task was learned during pretraining to extract the most useful representations for keyword spotting without the need of any labeled data.
    
    \item Then, a two-stage downstream task based on siamese neural networks  and PHOC attributes is used to further  promote the representation and makes it more powerful in  retrieving the best matching images  of a given query image.

    \item
    Extensive comparative experiments are achieved to validate the efficiency of our proposed method, involving three handwritten word images datasets.  We  demonstrate that our proposed method can be generalized across different databases and languages. We show also the effectiveness of our method to deal with data variability as well as data scarcity issues. 
      
\end{itemize}

The rest of this paper is organized as follows. In Section \ref{sec_related_work} we provide a review of prior works on keyword spotting for segmented documents. Then we introduce our proposed model in Section \ref{sec_proposed_method}. After that, experimental results and comparisons with existing methods will be described in Section \ref{sec_results}. Finally, in Section \ref{sec_conclusion} we draw the conclusions and we propose open challenges for future research directions.
 
\section{Related work}\label{sec_related_work}
Keyword spotting in handwritten documents has drawn the interest of the document analysis research community over the last  decades \cite{giotis2017survey} and it has still been challenging given the complexity of historical documents (diverse scripts, various writing styles, diverse noise, etc).
In the literature, there are several successful efforts that focus on different aspects of the word spotting issues like the data modality (printed, handwritten or scene text), method type (segmentation-based and segmentation free),  and the embedding (representation) type. In the following sub-sections, we categorize the related methods into two main families depending on the used representations: learning-free and learning-based techniques.  

\subsection{Learning-free representations}\label{learning_free_app}

Classical KWS methods were learning-free, different methods were built to find the best handcrafted matching features, or representations. Earlier works considered the handwritten word images as a temporal sequence to build a variable embedding representation. Most of these methods are based on profile features ~\cite{manmatha1996word} by computing each word image's column using diverse statistics at the pixel level. Within this scope, the Dynamic Time Warping (DTW) algorithm had been employed to match representations having variable lengths inspired by  its usefulness for sequence matching problems in speech. In \cite{rath2003word,zhang2003word}, the authors combined the DTW with the profile features (vertical profile, upper and lower word profile, and background to ink transitions) for better accuracy and  faster retrieval.  However, these simple structural features were leading to unsatisfactory accuracy.  Thus, statistical-based features, especially local gradient ones such as SIFT (Scale-invariant feature transform) \cite{lowe2004distinctive} and HOG (Histogram of oriented gradients) \cite{dalal2005histograms} were employed for word spotting. 

In \cite{rusinol2011browsing, rusinol2015efficient}, the authors propose a Bag Of Words (BOW) method which uses the SIFT descriptors to extract the local features, then projects them to a topic space that conserves the lexical content of the word images.  A similar approach with an addition of the corner detector features was also introduced   in \cite{shekhar2012word}.  Another approach that uses the projections of oriented gradients as descriptors was proposed in \cite{Retsinas2016}.  A sequence of
descriptors based on the combination of a zoning scheme and an appearance descriptor (or, modified Projections of Oriented Gradients) was also used in \cite{Retsinas2019} to represent the word images.  In \cite{Almazan2012,almazan2014segmentation}, document images were represented with a grid of HOG descriptors, then  the document regions that are most similar to a query word are located in a sliding-window  fashion. After that,   a second stage of re-ranking the best retrieved regions using a more discriminative BOW representation  was applied.   In \cite{Almazan2014}, a representation  to embed the word image to its corresponding text label in the same space using the developed PHOC was introduced. Additionally, recently, some graph based approaches have been introduced. In \cite{stauffer2018}, the text images  were represented with a graph structure, then  graph matching was applied for the spotting.

Despite the simplicity of the learning-free approaches and the ability to be adapted and applied in different domains, their results are still unsatisfactory. Thus, learning-based approaches have since been in vogue.

\subsection{Learning-based representations}\label{learning_based_app}

With the advances in machine learning, especially in  the deep learning field, features are now learned within the model instead of handcrafting them. 
Nowadays, machine learning based  word spotting models are  learning the features using labeled data.  Most of the developed approaches within this strategy are using CNNs. In \cite{jaderberg2014deep}, a CNN based model is employed to  detect words regions in natural images, then it recognizes these detected words. Another model called PHOCNet \cite{Sudholt2016} was proposed  to embed the image features extracted by the CNN layers into the PHOC representation. The representations are calculated  by applying a sigmoid on   the final fully connected layer. This latter work was extended in \cite{Sudholt2017,Sudholt2018} to improve the spotting accuracy. Another PHOC based model was introduced in \cite{Rusakov2018}, where instead of a distance-based matching of the retrieved words,  a probabilistic ranking was used for better performance.   In \cite{Krishnan2016}, a model called HWNet was introduced to match image collections containing handwritten content, where the representations were learned using a CNN. The HWNet representations were also used in \cite{Krishnan2016deep, Krishnan2018} by embedding them into the word attribute space. This was done by  training a classifier to  project both image and textual attributes to a common subspace.
In \cite{Serdouk2019}, a triplet loss based CNN approach was proposed to learn the representations by reducing the distance of the anchor image to a similar (positive) image while enlarging the distance to a different (negative) word image.

Other approaches were proposed based on graphs, for instance,  \cite{riba2021}  learned the representations by a Graph Neural Network  (GNN).  A message passing neural network was employed to capture the graph structure, that is used for the distance computation.

\subsection{Self-Supervised Learning in document processing/analysis}\label{sec_selfsupervised_app}

 Learning-based approaches are effective when a large amount of labeled data is available. However, with the continual growth of deep learning architectures, they start overfitting on the usual  datasets and requiring more samples, which is a challenging problem.   Thus, a  recent development in self-supervised learning \cite{carion2020end,bao2021beit,MAEVit}, especially with the rise of transformers architectures \cite{vaswani2017attention, dosovitskiy2021an}, is now appearing as a solution. Self-supervised methods aim to benefit from a huge amount of unlabeled data that can be added to the labeled datasets for training.  Nowadays, several document analysis papers are proposed following this strategy.  These approaches were developed for instance in document understanding in terms of classification, layout analysis or entity extraction  \cite{xu2020layoutlmv2, li2021selfdoc, li2022dit, appalaraju2021docformer} with different pretraining objectives (text-image matching, text-image alignment, masked visual-language modeling, document reconstruction, etc). These models were proposed for optical text images and using an OCR for the pretraining. Thus,  their learned representation can not be used for fine-tuning our model on handwritten KWS. Other approaches were developed for handwritten/scene text recognition by learning a representation that considers the text image as a sequence of characters in a sequence-to-sequence contrastive learning fashion. In \cite{aberdam2021sequence}, the authors first applied transformation on  each unlabeled word. Then, the feature maps are divided into different instances, over which the contrastive loss is computed, where from each image several positive pairs and multiple negative examples are extracted. However, the segmentation of words was not accurate due to the use of unlabeled data,  which makes the method learns a representation of a sequence of "word parts" rather than the actual characters. In \cite{zhang2022context},  a similar approach was proposed by concatenating different unlabeled words to produce two views, then their aligned features are used with a contrastive loss to pull together the positive samples and push apart the negative samples. However, concatenating unlabeled words can cause the consideration of positive words as negatives.  
 More recently, a self-supervised approach for HTR using the  masking-recovering strategy with the generative models  was proposed \cite{souibgui2022text}. The method applies different degradation on the unlabeled word images (masking, blurring and background noise) and then learns to reconstruct the original clean image as a pretraining task. 
 It was shown that this method overcomes the contrastive learning drawbacks of requiring large batch sizes and large data points while learning more effective and robust representation, especially for fine-tuning with fewer data. Another method that combines contrastive learning based pretraining with the masking based one was also proposed in \cite{yang2022reading}.

\section{Proposed method}\label{sec_proposed_method}
 
\begin{figure*}[!htb]
\centering
\includegraphics[width=120mm, height=100mm]{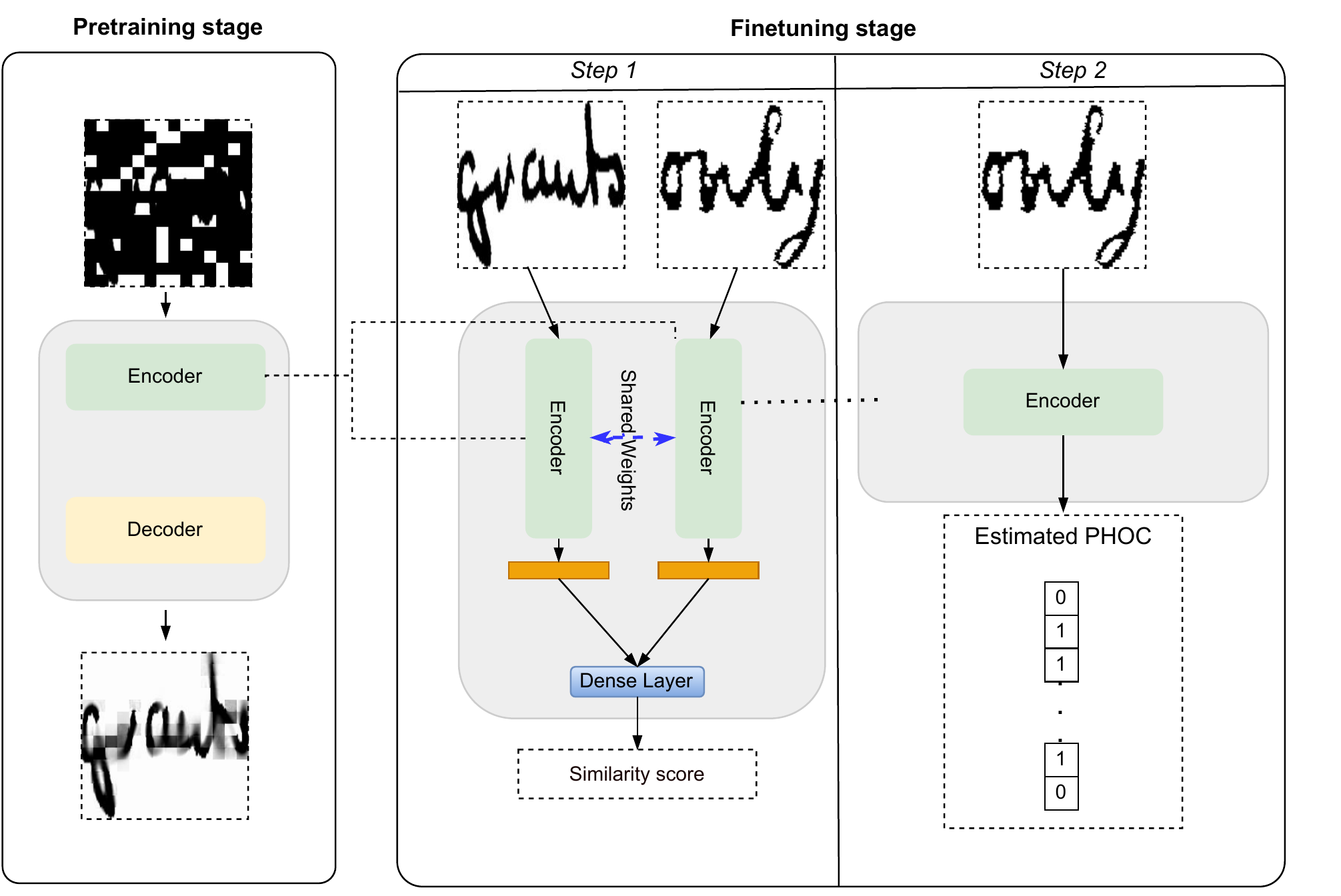}
\caption{Pipeline of the proposed \textit{ST-KeyS} framework. It consists of two stages: a pretraining stage and a fine-tuning stage.}
\label{fig:pipeline}
\end{figure*}

In this section, we present the proposed framework for word spotting in handwritten document images. We refer to this framework as \textit{ST-KeyS}: Self-supervised Transformer for Keyword Spotting. The idea is to benefit from the big amount of unlabeled data to learn word image representation in a self-supervised way. Then further improve and align these representations using the available labeled data.  Fig.~\ref{fig:pipeline} illustrates the overall proposed framework,  similar to other recent works in self-supervised representation learning, our proposed method consists of two phases: 
\begin{itemize}
    \item Pretraining phase: a self-supervised pretraining using a  masked vision transformer autoencoder is performed to extract useful representations from the unlabeled text images.
    
    \item Fine-tuning phase: a downstream task  based on siamese neural networks used to learn deep representations (step1) of the word image followed by PHOC embedding enabling the extraction of contextual information from the word (step 2).
\end{itemize}

\FloatBarrier
\FloatBarrier
In the following sub-sections, we first describe the used  masked autoencoder architecture for self-supervised word representation learning. Second, we describe the used model based on the SNN architecture and PHOC representations for the down-stream task.
\subsection{Pretraining phase: Learning deep representation using a masked autoencoder model  }\label{sec_MAE}

\begin{figure*}[!htb]
\centering
\includegraphics[width=\linewidth]{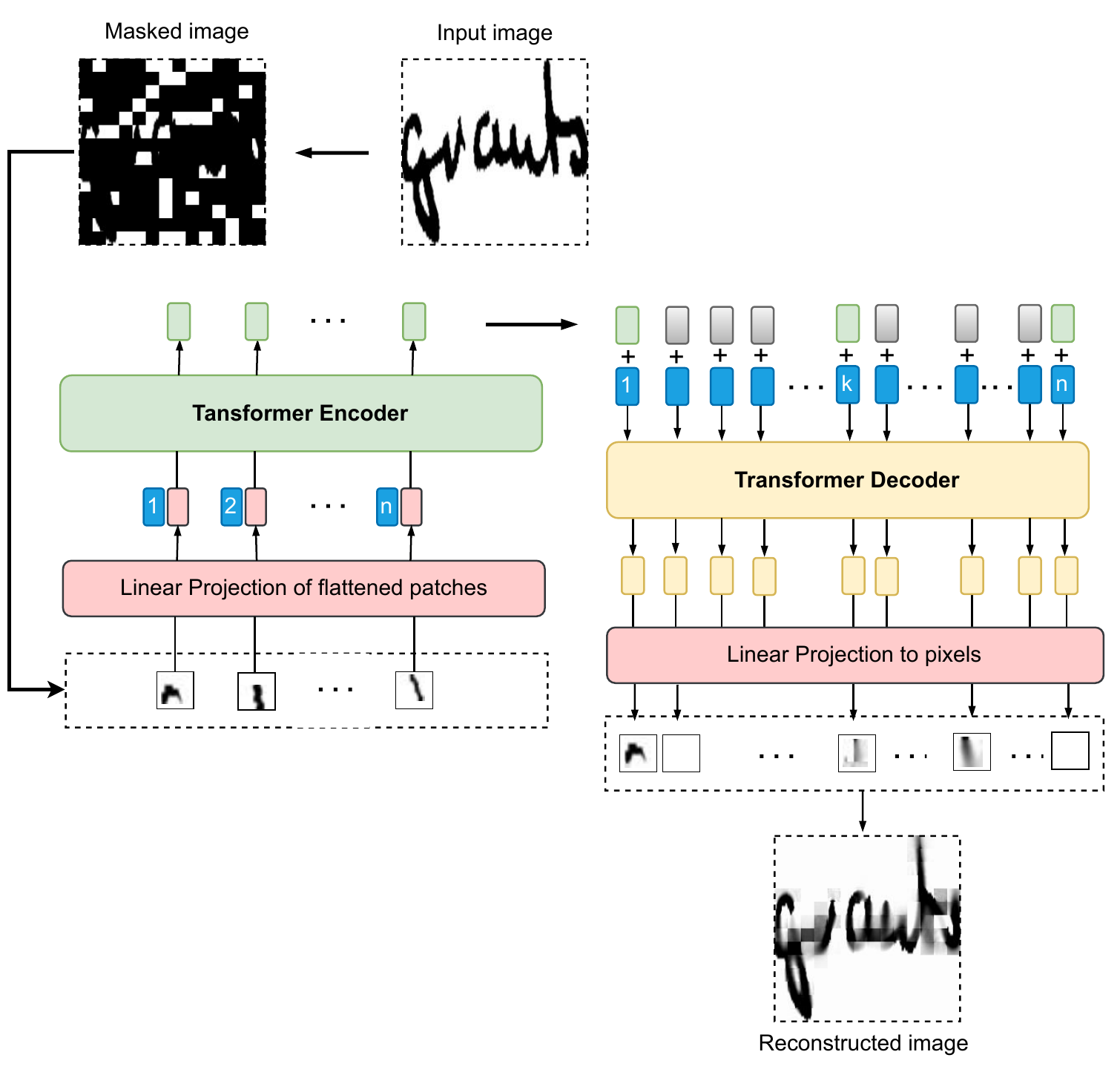}
\caption{Proposed masked encoder-decoder model for word representation learning.}
\label{fig:arch_MAE}
\end{figure*}

The pretraining phase is used to learn a word representation model in an unsupervised fashion.  Inspired by the work \cite{MAEVit}, we propose to use a masked autoencoder based on vision transformer. Fig.~\ref{fig:arch_MAE} illustrates the architecture of the used masked encoder-decoder. 
Same as the  ViT model \cite{dosovitskiy2021an}, each input image $x \in R^{H \times W \times C}$ ($H$, $W$ and $C$ denote respectively the height, width, and number of channels of $x$) is split into a set of $n$ non overlapping regular patches of size $p \times p \times C$. Then a fraction of 75\% of these patches are randomly masked according to a uniform distribution. The input image is then represented by only the remaining visible patches at the input of the encoder.

\FloatBarrier
\FloatBarrier

Different from conventional autoencoders, the used masked encoder-decoder operates in an asymmetric fashion \cite{MAEVit} enabling the encoder to operate on only the partial observed signal (without the masked patches) and the decoder to rebuild the full image based on the representation given by the encoder and the masked patches. As shown in Fig.~\ref{fig:arch_MAE}, the encoder takes as input the visible patches with their positional information and maps them into a latent representation, while the decoder reconstructs the masked pixels based on this latent representation.

\begin{figure*}[!htb]
\centering
\includegraphics[width=\linewidth]{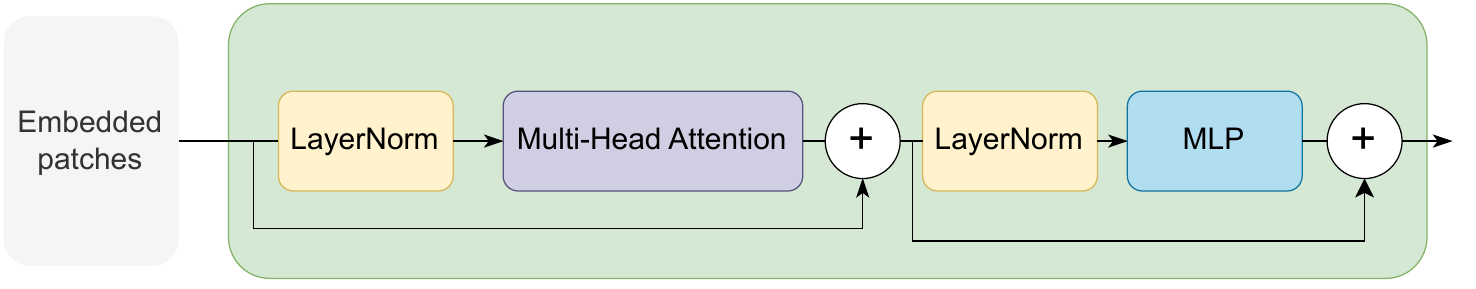}
\caption{Architecture of the Encoder used in our work.}
\label{fig:arch_encoder}
\end{figure*}

\noindent \textbf{The Encoder.}
The used encoder architecture is a vision transformer  \cite{dosovitskiy2021an} that operates only on visible patches. First, the visible patches are embedded through a linear projection operation into a patch embedding set (denoted $E_i$). In order to allow the model to capture the spatial structure of the 2D image, each patch embedding in $E_i$ is then associated with its positional information within the original image. The resulting embedding set $E'_i$ is then passed as input to the encoder. After that, a number of transformer blocks are employed to map this embedding $E'_i$ into a latent representation (output patch embedding set referred as  $E'_O$). 
These encoder's blocks have the same structure as in \cite{dosovitskiy2021an}. Each block consists of multi-head self-attention and multi-layer perceptron (MLP) alternating layers. 
The MLP network consists of a single layer with a Gaussian Error Linear Unit (GELU) activation function. Layer normalization (LayerNorm) is applied before each transformer block allowing to train efficiently deep encoders \cite{Wang2019}, and we used residual connections after every block. 
An overview of the encoder architecture used in our work is illustrated in Fig.~\ref{fig:arch_encoder}. 
It should be noted also that our encoder is applied only on visible set of patches. In our case, it operates on only a fraction of 25 \% of the full set of patches.

\FloatBarrier
\FloatBarrier

\noindent \textbf{The Decoder.}
The decoder is used to predict the pixel values for masked patches and then reconstruct the input image based on the latent representation provided by the encoder. The decoder takes as input the representation of the encoded patches $E'_O$ (visible patches) and the masked patches with their positional information which provides their arrangement location inside the image. 
Similar to the encoder architecture, the decoder has also series of transformer blocks. The decoder produces a vector of pixel values to represent each patch. All vectors are then reshaped to reconstruct the image. Similar to \cite{dosovitskiy2021an}, we use the mean squared error (MSE) between the masked patches from the original and reconstructed images as the loss function. 

We note that the decoder is only used during the pretraining phase in order to perform the image reconstruction and it will be discarded during the second phase. 

\subsection{Fine-tunig phase: Proposed down-stream task for word spotting}\label{sec_Siamese}

The goal of our study is to propose a spotting method dealing with the query by example scenario for KWS that can handle existing challenges in handwritten documents such as writing style variations.
For that, we introduce as the downstream task a KWS method based on a siamese neural network (SNN) architecture coupled with PHOC embedding. The SNN is a dedicated network having two inputs and it is able to learn relevant representations of handwritten images and therefore it allows distinguishing between similar and non-similar pairs of images. While the SNN network is used to learn a descriptor that is representing the image, the PHOC embedding is involved to provide a contextual representation of the characters embedded inside the word. Such process shall be a suitable way to benefit from the visual (image) and language (text) modalities when producing the final representations.

\subsubsection{Visual image representation}\label{sec_image_represention}
\begin{figure*}[!htb]
\centering
\includegraphics[width=\linewidth]{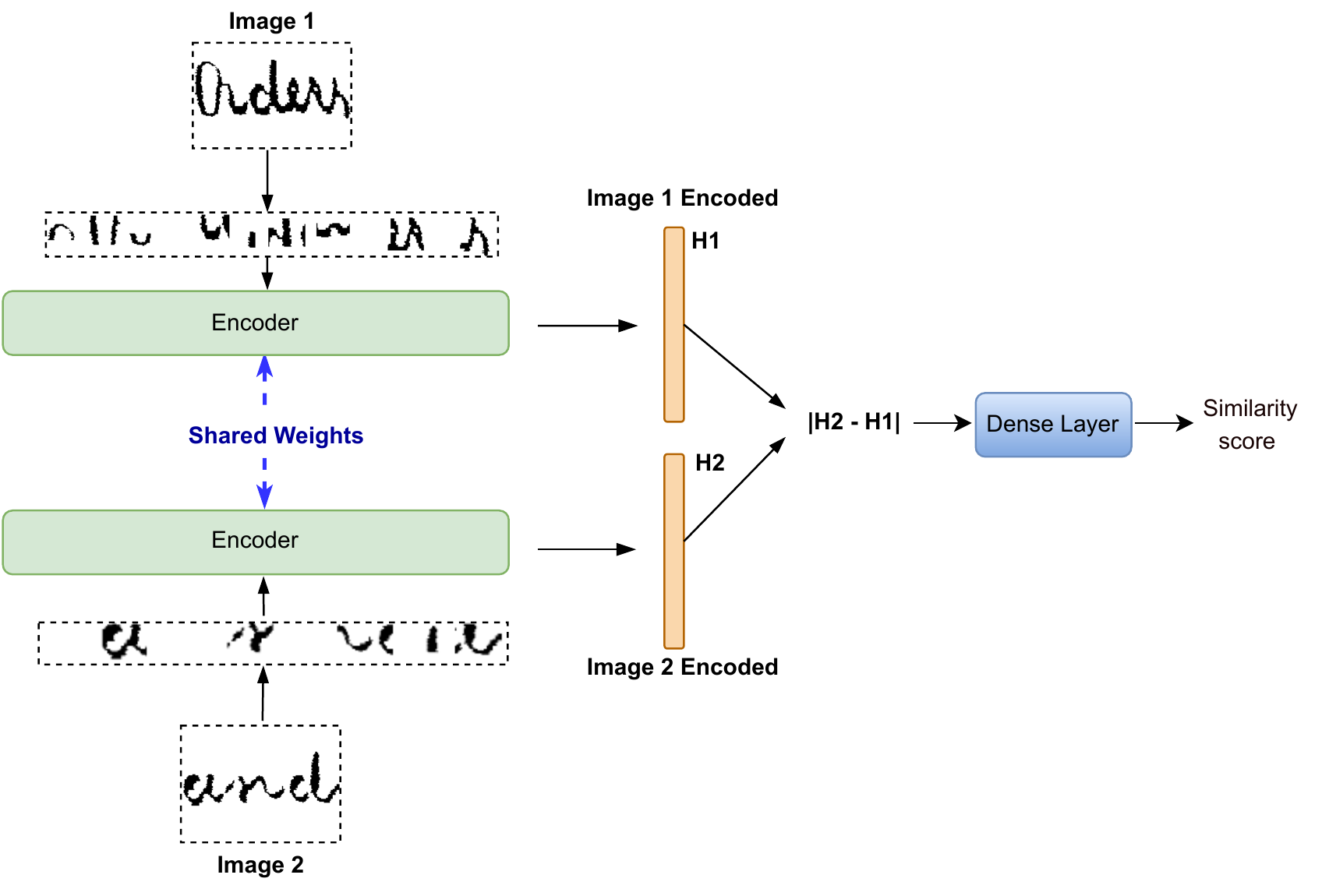}
\caption{The proposed SNN-transformer architecture. The H1 and H2 refer to numerical feature vectors corresponding to Image 1 and Image 2, respectively.}
\label{fig:arch_siam}
\end{figure*}

Once the masked autoencoder is pretrained in a self-supervised way on large unlabeled dataset, the decoder is discarded and only the encoder part is retained for the downstream task. Therefore, the pretrained encoder is  used as a backbone to build the SNN-transformer architecture as shown in Fig.~\ref{fig:arch_siam}.  The pretrained encoder is considered as a good starting point for the SNN model because it has learned word image representation from unlabeled data during the first pretraining phase without any human supervision or annotation. It should be noted that during this phase, the encoder is applied to uncorrupted (non-masked) images represented by the full set of their patches.  As shown in Fig.~\ref{fig:arch_siam}, the SNN-transformer architecture comprises two identical encoder blocks, with shared weights, followed by a dense layer. The dense layer involves a Linear layer with dropout and sigmoid as  activation function.

\FloatBarrier
\FloatBarrier
 Given two input images, the embedded features are propagated, each through an encoder block. The SNN-transformer works in parallel on the two different input vectors to compute comparable output vectors. These output vectors are compared using a distance function, corresponding to the absolute values of the descriptors subtraction operation, and then passed to the fully connected layer with a sigmoid activation function to return a similarity measure indicating whether or not the two images belong to the same class.
The learning process is designed to adjust the embedding to be representative and enables thereafter the SNN-transformer to extract relevant features as effectively as possible. 
\subsubsection{PHOC word embedding}\label{sec_PHOC}
Once the encoder has been fine-tuned firstly in a siamese fashion using a few labeled data, we use this latter as the main component for the second adaptation stage. This stage is performed to align the visually extracted features by the previous stage with the text based features. Particularly, we make use of the textual representation referred to as PHOC embedding. This representation combines the histogram of characters in several spatial regions in a pyramidal manner. Here, each feature indicates the presence or absence of a specific character in a given spatial region.

\FloatBarrier
\begin{figure*}[!htb]
\centering
\includegraphics[width=120mm,height=70mm]{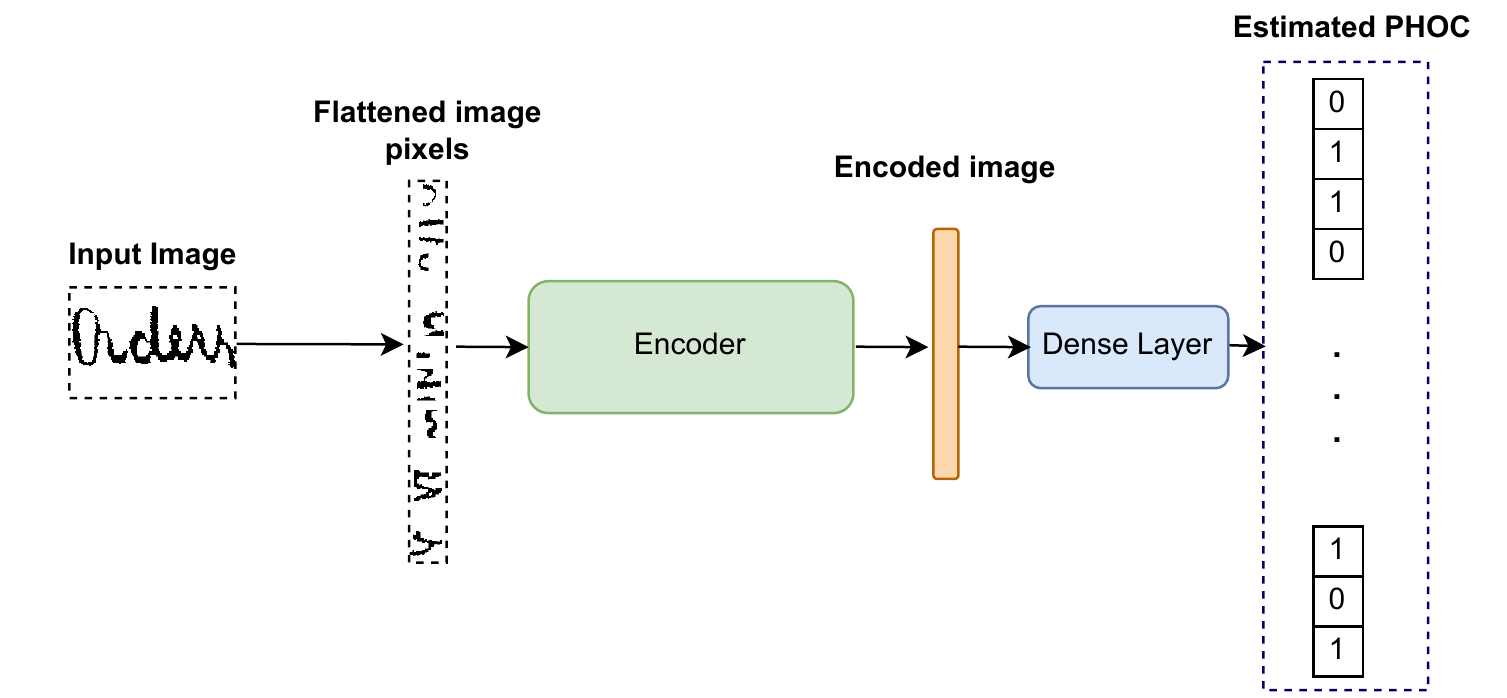}
\caption{The proposed PHOCNet-transformer architecture.}
\label{fig:arch_phoc}
\end{figure*}
\FloatBarrier

\begin{figure*}[!htb]
\centering
\includegraphics[width=110mm,height=40mm]{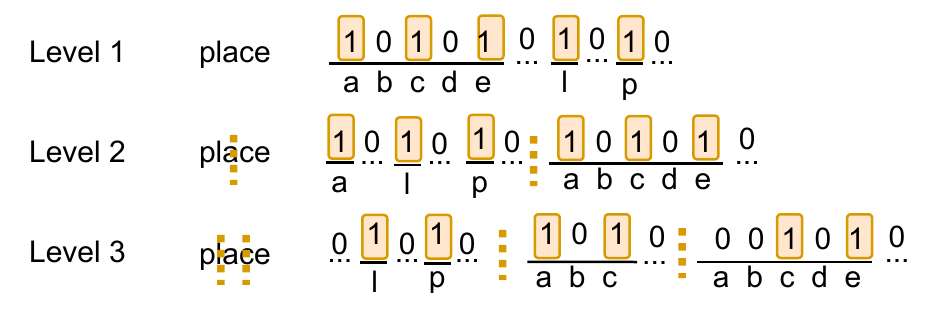}
\caption{Visualisation of the PHOC embedding extracted from a given text at level 1, 2 and 3.}
\label{fig:phoc_level}
\end{figure*}

The PHOCNet-transformer's pipeline is visualized in Fig.~\ref{fig:arch_phoc}. It consists of a single encoder block followed by a final dense layer. This last layer is a Linear layer with dropout and Sigmoid activation function. Similar to \cite{Sudholt2016}, we represent a word image using PHOC embedding with 2, 3, 4 and 5 levels. For instance, within the second level, the PHOC captures the presence of a certain character in the first or the second half of the word. Fig.~\ref{fig:phoc_level} provides an exemplary illustration of PHOC extraction from a given text string. This yields a binary histogram having a size of 504. Additionally, we make use of the 50 most frequent bigrams (level 2).  Thus, by the use of the Latin alphabet (lower case) with the ten digits, the resulting PHOC has a size of 604 which is corresponding to the output size of the defined dense layer.

Learning PHOC representations allows the adjustment of the embedding to be representative and enables the network to perform the spotting accurately. Therefore, after having trained our encoder using the PHOC representations, we make use of the layer preceding the final dense layer of this model to extract feature vectors representing the word images. The extracted feature vectors are then reduced to 400 dimensions using Principal Component Analysis (PCA), which is popularly used as a dimensionality reduction technique. To perform the word spotting task, we use the cosine distance metric which is calculated from a pair of feature vectors that each corresponds to a word image.

\FloatBarrier
\FloatBarrier

\section{Experiments}\label{sec_results}
In this section, we first describe the datasets used in our experiments. We then outline the implementation details that are most relevant to our approach. Lastly, we report our results and provide a comparison with state-of-the-art methods.

\subsection{Datasets}\label{databases}
Four databases of handwritten documents are used in our experiments: while the IAM off-line database is used to pretrain the model in a self-supervised way, three other datasets are used to fine-tune and evaluate the KWS proposed method. These three datasets are the Botany (BOT) dataset, Alvermann Konzilsprotokolle dataset (AK), and the George Washington dataset (GW) which are categorized as historical collections.

\begin{figure}[ht]
\centering
\begin{minipage}[t]{.45\textwidth}
    \begin{minipage}[t]{\textwidth}
    \centering
        \includegraphics[width=55mm,height=30mm]{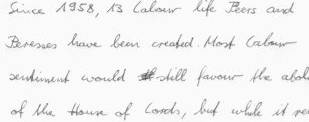}\vspace{-0.03\linewidth}
            \subcaption{IAM dataset\\ }\label{fig:1}
    \end{minipage}
    \vspace{5.5pt} \\
   \noindent
    \begin{minipage}[t]{\textwidth}
    \centering
        \includegraphics[width=55mm, height=30mm]{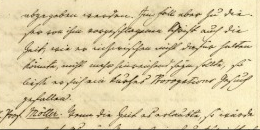}\vspace{-0.03\linewidth}
        \subcaption{AK dataset \\ }\label{fig:22}
    \end{minipage}
    \vspace{0.03\linewidth}
\end{minipage}
\begin{minipage}[t]{.45\textwidth}
    \begin{minipage}[t]{\textwidth}
    \centering
        \includegraphics[width=55mm, height=30mm]{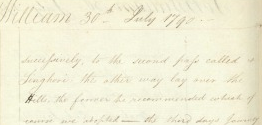}\vspace{-0.03\linewidth}
        \subcaption{BOT dataset \\ }\label{fig:11}
    \end{minipage}
    \vspace{5.5pt} \\
   \noindent
    \begin{minipage}[t]{\textwidth}
    \centering
        \includegraphics[width=55mm, height=30mm]{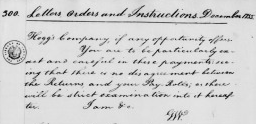}\vspace{-0.03\linewidth}
        \subcaption{GW dataset \\ }\label{fig:2}
    \end{minipage}
    \vspace{0.03\linewidth}
\end{minipage}
    \caption{Sample document images from the used datasets. The IAM database is used during the self-supervised pretraining phase. The BOT, AK and GW datasets are used for the second phase and thereby for word spotting evaluation experiments. We notice that BOT, AK and GW datasets are categorized into historical collection.}
    \label{fig:used_dataset}
\end{figure}

\noindent \textbf{IAM dataset.} 
The IAM dataset  \cite{IAM}
involves forms of handwritten English text useful for training and testing handwriting recognition models and writer identification methods. IAM contains handwriting belonging to 657 different writers which lead to a large variety of writing styles. This database consists of 1.539 pages of scanned text from the Lancaster–Oslo/Bergen corpus \cite{FISCHER2012934}, leading to 115.320 isolated and labeled word images split into training, validation and test partitions. Due to its large style variability, we have used the IAM dataset for the self-supervised pretraining phase. A sample document image is shown in Fig.~\ref{fig:1}.

\noindent \textbf{Botany dataset (BOT).} This dataset is a part of ICFHR 2016 Handwritten Keyword Spotting Competition \cite{Pratikakis2016}. It consists of more than 100 different botanical documents made between the year 1800 and 1850 by the then government of British India. 
These documents are written in English and show some signs of deterioration, especially fading as shown in Fig.~\ref{fig:11}.  Variations in writing style are noticeable, especially in scaling and intra-word variations.

\noindent \textbf{Alvermann Konzilsprotokolle dataset (AK).} 
This dataset is also a part of ICFHR 2016 Handwritten Keyword Spotting Competition \cite{Pratikakis2016}. It is a historical document collection consisting of 18.000 pages of handwritten minutes of official meetings in the central administration of the University of Greifswald between the year 1794 and 1797. These documents are written in German and show minor signs of degradation as shown in Fig.~\ref{fig:22}. This dataset presents low variation in the writing style.

\noindent \textbf{George Washington dataset (GW).} The GW dataset is well known to the community of word spotting. It consists of 20 pages of letters written by George Washington and his secretaries in 1755. 
It contains 4.860 annotated word images. Due to the small size of the dataset, there are no available standard partitions or query selection for GW. Thus, a four-fold cross validation setup is generally adopted by the majority of learning-based recent works. We have adopted this setup using the same split as in \cite{Almazan2014}. 
For each test split, words having at least two instances are selected as query images in a leave-one-out fashion.  
A sample document image from the GW database is shown in Fig.~\ref{fig:2}.

\subsection{Evaluation protocol}\label{protocol}

For the word spotting task, the aim is to retrieve all instances of the query words in a dataset's partition. Given a query image, the database items are sorted with respect to their similarity to the introduced query following the protocol presented in \cite{Almazan2014}. In our work, we use all the words of the test set as a query in a leave-one-out style. When performing keyword spotting, the query image is removed from the dataset, and queries that have no extra relevant words in the database are discarded. However, the remaining irrelevant words are maintained in the test set to act as distractors in the ranking process. This protocol is adopted in all our experiments using the three evaluation sets, namely, BOT, AK and GW databases. Particularly, for the GW dataset, we evaluate our proposed method in a four-cross validation fashion to enable comparison with state-of-the-art systems. 
For the BOT and AK datasets, there were three partitions of training sets: small, medium, and large. Here we use the smallest partition for conducting experiments. 
Table~\ref{tab:stats_datasets} provides statistics on the databases used in our experiments. To assess the performance of the performed method, we use the mean Average Precision (mAP) as the evaluation measure.

\begin{table}[ht]
\caption{{Statistics derived from the datasets used for evaluation.
}}
\centering
\begin{tabular}{|p{6cm}|p{1.5cm}|p{1.5cm}|p{1.5cm}|}
\hline
Dataset & Keywords  & Test & Train \\ \hline
Botany (BOT)  & 150     & 3380 & 1684  \\  
Alvermann Konzilsprotokolle (AK)      & 200     & 3737  & 1849 \\ 
George Washington (GW)     & 901     & 1215 & 2447  \\  \hline
\end{tabular}
\label{tab:stats_datasets}
\end{table}

\subsection{Implementation details}\label{sec_implementation_details}

Since the proposed method is based on the masked encoder-decoder architecture, we initially performed several preliminary experiments to select the best model architecture leading to the best results. Moreover, we have picked up some parameters that have been proven successful in \cite{MAEVit}. Subsequently, we trained the network in an end-to-end fashion. Table.~\ref{tab:params} describes the network's architecture and the parameter setting.
 
\begin{table}[ht]
\caption{{Proposed model's hyperparameters used in our work.}}
\centering
\begin{tabular}{|p{5.5cm}|p{3cm}|}
\hline
Parameter           & Value   \\ \hline
Patch size          & $16\times16$      \\  
Image size          & 224x224 \\  
Train size\ding{62}          & 60k     \\  
Encoder embedding dimension & 768     \\  
Encoder layers number      & 12      \\  
Encoder head number       & 12      \\ 
Decoder embedding dimension   & 384     \\ 
Decoder layers number     & 4       \\ 
Decoder head number & 6       \\ \hline
\end{tabular}

\ding{62} \textit{Training set is augmented using different image transformation operations.}
\label{tab:params}
\end{table}

The proposed framework was developed using the Pytorch library designed for deep learning. To  pretrain the self-supervised model, we used a maximum of 500 epochs, with a batch size of 20. During the pretraining stage we used AdamW, an improved version of the algorithm \cite{adam2014}, for parameter optimization. The learning rate is initially set to $1.5\mathrm{e}{-4}$  and decreased using a scheduler.
For the fine-tuning stage, we employed the SGD (Stochastic gradient descent) optimization algorithm with a learning schedule. An initial learning rate of 0.01 was used when fine-tuning the siamese neural network. This rate is declined every 3 epochs. The experiments were performed using an NVIDIA TITAN Xp GPU. To improve the generalization power of the proposed \textit{ST-KeyS} model, we used different random transformation operations, as data augmentation techniques, such as erosion, dilation, and skewing.
\subsection{KeyWord Spotting Results}\label{sec_results_discussion}
We present in this section the obtained results. It should be noted that in all our experiments, we rely on the Cosine similarity distance as a  metric allowing us to measure the similarity between two feature vectors. This metric computes the angle's cosine between two vectors and identifies if two vectors point approximately in the same direction.

\subsubsection{Evaluation on ICFHR 2016 competition's datasets}\label{sec_results_ICFHR16}

The evaluation process is conducted firstly on the \textit{ICFHR 2016} competition's datasets, namely, the BOT and the AK.  Table \ref{tab:resuts_botany_AK} shows the obtained results.  
We recall that the \textit{ST-KeyS} model is first pretrained on the IAM database in a self-supervised way in order to learn deep representation from word images. Secondly, the pretrained encoder is fine-tuned using a few labeled data from the target dataset. 
Since the IAM, BOT, AK and GW databases are belonging to different datasets distributions, we are intended to deal with challenges related to domain variations and writers styles shifts. 
That's why we extend the pretraining stage with some epochs on a subset from the target dataset (BOT, AK and GW) in a self-supervised way (without annotation/labels). Such a process allows to reduce the gap between domains originating from the calligraphic style fluctuations of the IAM and the used evaluation datasets.

From Table~\ref{tab:resuts_botany_AK}, we  notice that the proposed \textit{ST-KeyS} model yields good results on the BOT dataset reaching a performance of 59.10\% which outperforms TPP-PHOCNet(CPS), PHOCNet(CPS) and Triplet CNN architectures \cite{Sudholt2018}. However, we obtained moderate results on the AK dataset comparing to TPP-PHOCNet(CPS) and PHOCNet(CPS) architectures \cite{Sudholt2018}.
Moreover, we observe that the \textit{ST-KeyS} leads to better results than learning-free based methods \cite{Retsinas2016,stauffer2018,Retsinas2019} on both the BOT and AK datasets although they represent different calligraphic styles compared to the IAM database utilized in the first pretraining stage and also despite the difference in the written language script.
Additionally, our \textit{ST-KeyS} model gives better results compared to the Graph Edit Distance method proposed in \cite{riba2021} thanks to the transformer's capacity to focus on meaningful information included in the word image. This capacity is owed to the attention mechanism allowing to extract discriminant features from word images and which are used later in the matching process.

In summary, we note that the \textit{ST-KeyS} semi-supervised model outperforms the fully supervised PHOC-based methods \cite{Sudholt2018} that were conducted in the literature on the BOT dataset and yields moderate results with these methods on the AK dataset. 
Generally, these methods rely on synthetically generated data to learn the PHOC contextual representations. Thus, such models are able to treat large vocabulary allowing to alleviate to the out of vocabulary words issue \cite{jemni2019out}.
In the remainder of our work, we limit the PHOC learning stage to the data provided in the competition image databases. In other words, we are based on a closed vocabulary since our goal is to build a robust model allowing to retrieve handwritten words independently of the out of vocabulary word problem. Therefore, it is unfair to compare our results to PHOC-based methods that were conducted in the literature. For that reason, we restrict the comparison to methods using closed vocabulary. 

\FloatBarrier
\begin{table}[ht]
\caption{{Evaluation results (mAP \%) of the word spotting proposed method performed on the BOT and the AK datasets.}}
\centering
\begin{tabular}{|p{7cm}|p{1.5cm}|p{1.5cm}|}
\hline
Method                 & BOT         & AK            \\ \hline \hline
\multicolumn{3}{|l|}{\textit{Learning-free methods}} \\ \hline
Projections of Oriented Gradients ~\cite{Retsinas2016}&    46.70    &   56.50 \\
Graph matching ~\cite{stauffer2018} & 49.57         & 77.24          \\
PSeq-mPOG+MISM ~\cite{Retsinas2019}  & 58.30  & 76.20 \\ 
 \hline\hline
\multicolumn{3}{|l|}{\textit{Learning-based methods}} \\ \hline 
Graph Edit Distance ~\cite{riba2021}       & 52.83          & 79.55         \\ 
TPP-PHOCNet(CPS) ~\cite{Sudholt2018}\ding{61}\ding{71}      & 51.25          & 90.97  \\ 
PHOCNet(CPS)~\cite{Sudholt2018}\ding{61}\ding{71}   & 45.82          & 88.31  \\
Triplet-CNN ~\cite{Sudholt2018}\ding{61}\ding{71}   & 54.95          & 82.15  \\
\hline\hline 
\textbf{Ours (\textit{ST-KeyS} })\ding{61}      & \textbf{59.10} & \textbf{85.16} \\ \hline 
\end{tabular}

\ding{61} \textit{Methods based on PHOC representation.} 

\ding{71} \textit{Methods based on generated synthetic data (large vocabulary).}  
\label{tab:resuts_botany_AK}
\end{table}
\FloatBarrier

\subsubsection{Evaluation on the GW dataset}\label{sec_results_GW}

Further experiments are conducted on the GW dataset to assess the performance of our proposed \textit{ST-KeyS} method. The obtained results on GW dataset are illustrated in Table \ref{tab:resuts_GW_comparaison}. \textit{ST-KeyS} achieved a mAP of 95.70\% on the GW database. Table \ref{tab:resuts_GW_comparaison} shows that our proposed method leads to interesting results that are exceeding all considered learning-free methods \cite{Almazan2012,Almazan2014,Retsinas2016,Retsinas2019}. This result proves the efficiency of the \textit{ST-KeyS} model to extract deep meaningful features from word images enabling an accurate word spotting task compared to the use of handcrafted ones (HOG, mPOG, etc.). 
In addition, the \textit{ST-KeyS} method gives better results than the current state-of-the-art methods that are based on supervised approaches \cite{Serdouk2019,Sudholt2016,Sfikas2016} and using a closed vocabulary. Moreover, \textit{ST-KeyS} outperforms the graph-based approach \cite{riba2021} by a large margin.

It is noteworthy that our method focus on designing a robust model allowing retrieval of word images in handwritten documents when only few labelled data are available. Despite this, we can see from Table \ref{tab:resuts_GW_comparaison} that \textit{ST-KeyS} achieved competitive results compared to PHOC based methods trained on large generated synthetic data. \textit{ST-KeyS} outperforms the methods presented in \cite{Almazan2014,Krishnan2016} and  gets a slightly lower performance compared the methods \cite{Krishnan2019,Sudholt2017}  without the need for any external data.
 
\FloatBarrier
\begin{table}[ht]
\caption{{Evaluation results (mAP \%) of the word spotting proposed method performed
on the GW dataset.}}
\centering
\begin{tabular}{|p{8cm}|c|c|}
\hline
\textbf{Method}     &    \textbf{mAP \%}  \\ \hline \hline
\multicolumn{2}{|l|}{\textit{Learning-free methods}} \\ \hline
HOG + SVM  ~\cite{Almazan2012}  &  49.40   \\ 
DTW ~\cite{Almazan2014}  & 60.63  \\                              
FV ~\cite{Almazan2014}&   62.72  \\                               
Projections of Oriented Gradients ~\cite{Retsinas2016}&    37.00    \\
PSeq-mPOG+MISM ~\cite{Retsinas2019}    & 77.10   \\  
\hline\hline
\multicolumn{2}{|l|}{\textit{Learning-based methods}} \\ \hline
Zoning Aggregated features (CNN)  ~\cite{Sfikas2016}  & 58.30   \\ 
Softmax CNN  ~\cite{Sudholt2016} &    78.24  \\                   
Siamese Triplet CNN  ~\cite{Serdouk2019}&  91.63   \\ 
Graph Edit Distance  ~\cite{riba2021}   & 78.48 \\                
Att. + Platts (PHOC) ~\cite{Almazan2014}\ding{61}\ding{71}   & 93.04  \\ 
PHOCNet ~\cite{Sudholt2016} \ding{61}\ding{71}&   96.71 \\ 
TPP-PHOCNet  ~\cite{Sudholt2017}\ding{61}\ding{71}  & 97.78  \\   
HWNet ~\cite{Krishnan2016}\ding{61}\ding{71}  & 94.84 \\          
HWNet v2 (ROI) ~\cite{Krishnan2019}\ding{61}\ding{71}&   96.01 \\      
HWNet v2 (TPP) ~\cite{Krishnan2019}\ding{61}\ding{71}&   98.24 \\      

\hline\hline
\textbf{Ours (\textit{ST-KeyS})}\ding{61}    &  \textbf{95.70}   \\  
\hline 
\end{tabular}

\ding{61} \textit{Methods based on PHOC representations.}

\ding{71} \textit{Methods based on generated synthetic data (large vocabulary).}
\label{tab:resuts_GW_comparaison}
\end{table}
\FloatBarrier

\begin{figure*}[ht]
\begin{center} 
\centering
\begin{tabular}{p{1.8cm} p{1.5cm}p{1.5cm} p{1.5cm} p{1.5cm} p{1.5cm}}
\hline
\multicolumn{6}{l}{\centering Query  \qquad \qquad \qquad \qquad Top 5 retrieved images} \\ \hline

\\
\tcbox[sharp corners,colback=white, colframe=black,  boxsep=0pt, left=0pt, right=0pt,top=0pt,bottom=0pt, boxrule=0.5pt] {\includegraphics[width=15mm,height=8mm]{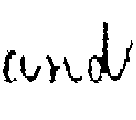}}   & \tcbox[sharp corners,colback=white, colframe=green,  boxsep=0pt, left=0pt, right=0pt,top=0pt,bottom=0pt, boxrule=0.5pt] {\includegraphics[width=15mm,height=8mm]{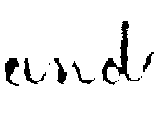}}   & 
\tcbox[sharp corners,colback=white, colframe=green,  boxsep=0pt, left=0pt, right=0pt,top=0pt,bottom=0pt, boxrule=0.5pt] {\includegraphics[width=15mm,height=8mm]{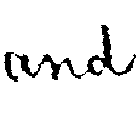}}  &
\tcbox[sharp corners,colback=white, colframe=green,  boxsep=0pt, left=0pt, right=0pt,top=0pt,bottom=0pt, boxrule=0.5pt] {\includegraphics[width=15mm,height=8mm]{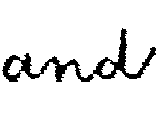}} &
\tcbox[sharp corners,colback=white, colframe=green,  boxsep=0pt, left=0pt, right=0pt,top=0pt,bottom=0pt, boxrule=0.5pt] {\includegraphics[width=15mm,height=8mm]{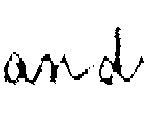}}&
\tcbox[sharp corners,colback=white, colframe=green,  boxsep=0pt, left=0pt, right=0pt,top=0pt,bottom=0pt, boxrule=0.5pt] {\includegraphics[width=15mm,height=8mm]{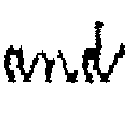}}

\\

\tcbox[sharp corners,colback=white, colframe=black,  boxsep=0pt, left=0pt, right=0pt,top=0pt,bottom=0pt, boxrule=0.5pt] {\includegraphics[width=15mm,height=8mm]{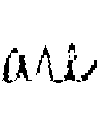}}   & \tcbox[sharp corners,colback=white, colframe=green,  boxsep=0pt, left=0pt, right=0pt,top=0pt,bottom=0pt, boxrule=0.5pt] {\includegraphics[width=15mm,height=8mm]{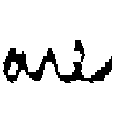}}   & 
\tcbox[sharp corners,colback=white, colframe=green,  boxsep=0pt, left=0pt, right=0pt,top=0pt,bottom=0pt, boxrule=0.5pt] {\includegraphics[width=15mm,height=8mm]{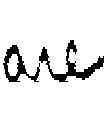}}  &
\tcbox[sharp corners,colback=white, colframe=green,  boxsep=0pt, left=0pt, right=0pt,top=0pt,bottom=0pt, boxrule=0.5pt] {\includegraphics[width=15mm,height=8mm]{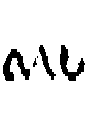}} &
\tcbox[sharp corners,colback=white, colframe=green,  boxsep=0pt, left=0pt, right=0pt,top=0pt,bottom=0pt, boxrule=0.5pt] {\includegraphics[width=15mm,height=8mm]{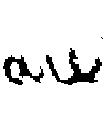}}&
\tcbox[sharp corners,colback=white, colframe=green,  boxsep=0pt, left=0pt, right=0pt,top=0pt,bottom=0pt, boxrule=0.5pt] {\includegraphics[width=15mm,height=8mm]{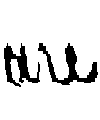}}

\\
\tcbox[sharp corners,colback=white, colframe=black,  boxsep=0pt, left=0pt, right=0pt,top=0pt,bottom=0pt, boxrule=0.5pt] {\includegraphics[width=15mm,height=8mm]{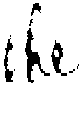}}   & \tcbox[sharp corners,colback=white, colframe=green,  boxsep=0pt, left=0pt, right=0pt,top=0pt,bottom=0pt, boxrule=0.5pt] {\includegraphics[width=15mm,height=8mm]{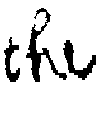}}   & 
\tcbox[sharp corners,colback=white, colframe=green,  boxsep=0pt, left=0pt, right=0pt,top=0pt,bottom=0pt, boxrule=0.5pt] {\includegraphics[width=15mm,height=8mm]{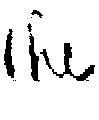}}  &
\tcbox[sharp corners,colback=white, colframe=green,  boxsep=0pt, left=0pt, right=0pt,top=0pt,bottom=0pt, boxrule=0.5pt] {\includegraphics[width=15mm,height=8mm]{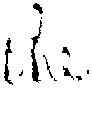}} &
\tcbox[sharp corners,colback=white, colframe=green,  boxsep=0pt, left=0pt, right=0pt,top=0pt,bottom=0pt, boxrule=0.5pt] {\includegraphics[width=15mm,height=8mm]{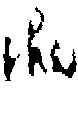}}&
\tcbox[sharp corners,colback=white, colframe=green,  boxsep=0pt, left=0pt, right=0pt,top=0pt,bottom=0pt, boxrule=0.5pt] {\includegraphics[width=15mm,height=8mm]{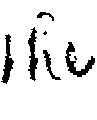}}
\\ 

\tcbox[sharp corners,colback=white, colframe=black,  boxsep=0pt, left=0pt, right=0pt,top=0pt,bottom=0pt, boxrule=0.5pt] {\includegraphics[width=15mm,height=8mm]{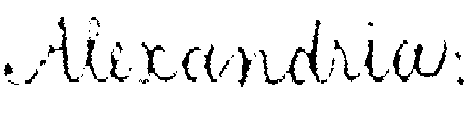}}   & \tcbox[sharp corners,colback=white, colframe=green,  boxsep=0pt, left=0pt, right=0pt,top=0pt,bottom=0pt, boxrule=0.5pt] {\includegraphics[width=15mm,height=8mm]{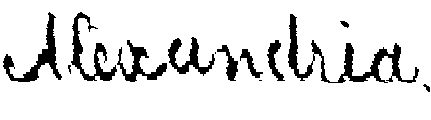}}   & 
\tcbox[sharp corners,colback=white, colframe=green,  boxsep=0pt, left=0pt, right=0pt,top=0pt,bottom=0pt, boxrule=0.5pt] {\includegraphics[width=15mm,height=8mm]{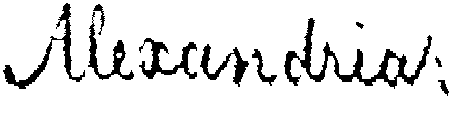}}  &
\tcbox[sharp corners,colback=white, colframe=green,  boxsep=0pt, left=0pt, right=0pt,top=0pt,bottom=0pt, boxrule=0.5pt] {\includegraphics[width=15mm,height=8mm]{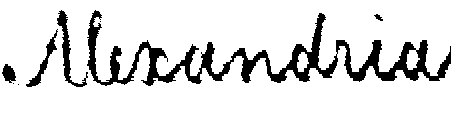}} &
\tcbox[sharp corners,colback=white, colframe=green,  boxsep=0pt, left=0pt, right=0pt,top=0pt,bottom=0pt, boxrule=0.5pt] {\includegraphics[width=15mm,height=8mm]{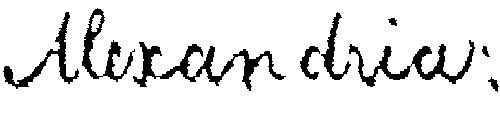}}&
\tcbox[sharp corners,colback=white, colframe=green,  boxsep=0pt, left=0pt, right=0pt,top=0pt,bottom=0pt, boxrule=0.5pt] {\includegraphics[width=15mm,height=8mm]{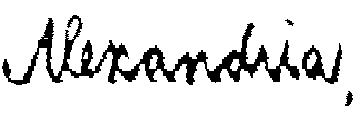}}
\\

\tcbox[sharp corners,colback=white, colframe=black,  boxsep=0pt, left=0pt, right=0pt,top=0pt,bottom=0pt, boxrule=0.5pt] {\includegraphics[width=15mm,height=8mm]{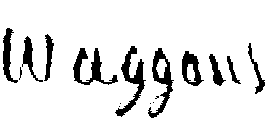}}   & \tcbox[sharp corners,colback=white, colframe=red,  boxsep=0pt, left=0pt, right=0pt,top=0pt,bottom=0pt, boxrule=0.5pt] {\includegraphics[width=15mm,height=8mm]{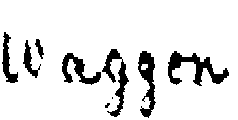}}   & 
\tcbox[sharp corners,colback=white, colframe=green,  boxsep=0pt, left=0pt, right=0pt,top=0pt,bottom=0pt, boxrule=0.5pt] {\includegraphics[width=15mm,height=8mm]{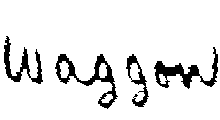}}  &
\tcbox[sharp corners,colback=white, colframe=green,  boxsep=0pt, left=0pt, right=0pt,top=0pt,bottom=0pt, boxrule=0.5pt] {\includegraphics[width=15mm,height=8mm]{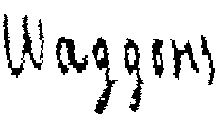}} &
\tcbox[sharp corners,colback=white, colframe=green,  boxsep=0pt, left=0pt, right=0pt,top=0pt,bottom=0pt, boxrule=0.5pt] {\includegraphics[width=15mm,height=8mm]{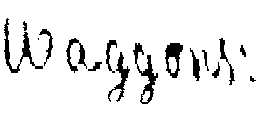}}&
\tcbox[sharp corners,colback=white, colframe=green,  boxsep=0pt, left=0pt, right=0pt,top=0pt,bottom=0pt, boxrule=0.5pt] {\includegraphics[width=15mm,height=8mm]{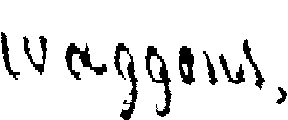}}
\\

\tcbox[sharp corners,colback=white, colframe=black,  boxsep=0pt, left=0pt, right=0pt,top=0pt,bottom=0pt, boxrule=0.5pt] {\includegraphics[width=15mm,height=8mm]{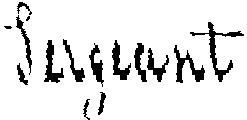}}   & \tcbox[sharp corners,colback=white, colframe=green,  boxsep=0pt, left=0pt, right=0pt,top=0pt,bottom=0pt, boxrule=0.5pt] {\includegraphics[width=15mm,height=8mm]{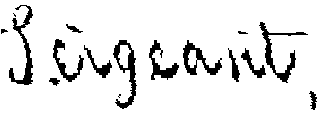}}   & 
\tcbox[sharp corners,colback=white, colframe=green,  boxsep=0pt, left=0pt, right=0pt,top=0pt,bottom=0pt, boxrule=0.5pt] {\includegraphics[width=15mm,height=8mm]{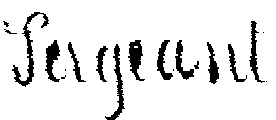}}  &
\tcbox[sharp corners,colback=white, colframe=red,  boxsep=0pt, left=0pt, right=0pt,top=0pt,bottom=0pt, boxrule=0.5pt] {\includegraphics[width=15mm,height=8mm]{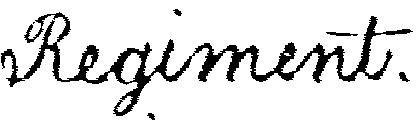}} &
\tcbox[sharp corners,colback=white, colframe=red,  boxsep=0pt, left=0pt, right=0pt,top=0pt,bottom=0pt, boxrule=0.5pt] {\includegraphics[width=15mm,height=8mm]{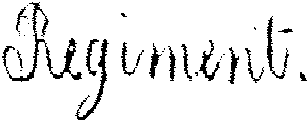}}&
\tcbox[sharp corners,colback=white, colframe=red,  boxsep=0pt, left=0pt, right=0pt,top=0pt,bottom=0pt, boxrule=0.5pt] {\includegraphics[width=15mm,height=8mm]{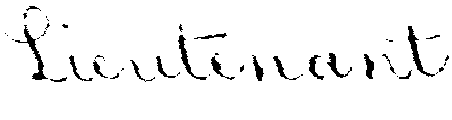}}
\\

\tcbox[sharp corners,colback=white, colframe=black,  boxsep=0pt, left=0pt, right=0pt,top=0pt,bottom=0pt, boxrule=0.5pt] {\includegraphics[width=15mm,height=8mm]{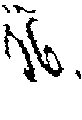}}   & \tcbox[sharp corners,colback=white, colframe=red,  boxsep=0pt, left=0pt, right=0pt,top=0pt,bottom=0pt, boxrule=0.5pt] {\includegraphics[width=15mm,height=8mm]{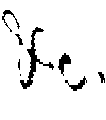}}   & 
\tcbox[sharp corners,colback=white, colframe=red,  boxsep=0pt, left=0pt, right=0pt,top=0pt,bottom=0pt, boxrule=0.5pt] {\includegraphics[width=15mm,height=8mm]{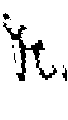}}  &
\tcbox[sharp corners,colback=white, colframe=red,  boxsep=0pt, left=0pt, right=0pt,top=0pt,bottom=0pt, boxrule=0.5pt] {\includegraphics[width=15mm,height=8mm]{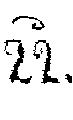}} &
\tcbox[sharp corners,colback=white, colframe=red,  boxsep=0pt, left=0pt, right=0pt,top=0pt,bottom=0pt, boxrule=0.5pt] {\includegraphics[width=15mm,height=8mm]{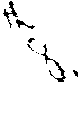}}&
\tcbox[sharp corners,colback=white, colframe=green,  boxsep=0pt, left=0pt, right=0pt,top=0pt,bottom=0pt, boxrule=0.5pt] {\includegraphics[width=15mm,height=8mm]{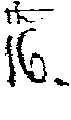}}
\\

\tcbox[sharp corners,colback=white, colframe=black,  boxsep=0pt, left=0pt, right=0pt,top=0pt,bottom=0pt, boxrule=0.5pt] {\includegraphics[width=15mm,height=8mm]{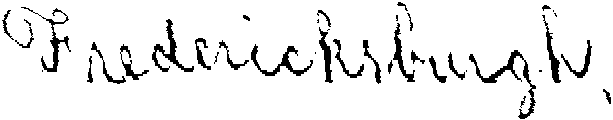}}   & \tcbox[sharp corners,colback=white, colframe=red,  boxsep=0pt, left=0pt, right=0pt,top=0pt,bottom=0pt, boxrule=0.5pt] {\includegraphics[width=15mm,height=8mm]{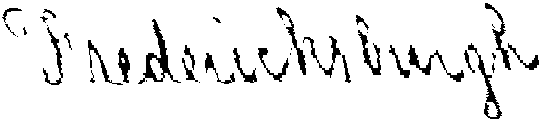}}   & 
\tcbox[sharp corners,colback=white, colframe=green,  boxsep=0pt, left=0pt, right=0pt,top=0pt,bottom=0pt, boxrule=0.5pt] {\includegraphics[width=15mm,height=8mm]{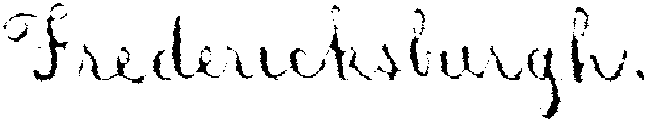}}  &
\tcbox[sharp corners,colback=white, colframe=red,  boxsep=0pt, left=0pt, right=0pt,top=0pt,bottom=0pt, boxrule=0.5pt] {\includegraphics[width=15mm,height=8mm]{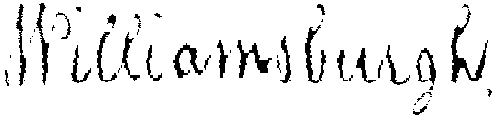}} &
\tcbox[sharp corners,colback=white, colframe=red,  boxsep=0pt, left=0pt, right=0pt,top=0pt,bottom=0pt, boxrule=0.5pt] {\includegraphics[width=15mm,height=8mm]{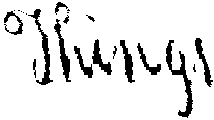}}&
\tcbox[sharp corners,colback=white, colframe=red,  boxsep=0pt, left=0pt, right=0pt,top=0pt,bottom=0pt, boxrule=0.5pt] {\includegraphics[width=15mm,height=8mm]{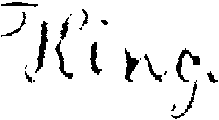}}
\\
\hline
\end{tabular}
\caption{Qualitative results of the proposed \textit{ST-KeyS} method on sample examples from the GW evaluation dataset. Green boxes correspond to correctly retrieved words and red boxes correspond to incorrectly retrieved words. }
    \label{fig:qualitative_result}
    \end{center}
\end{figure*}

Fig.\ref{fig:qualitative_result} shows qualitative evaluation of our proposed \textit{ST-KeyS} method on the GW database where the top five retrieved images for each query word image are given. 
This figure highlights the robustness of the features extracted using the proposed method. 
Such features make the method invariant across different word capitalization forms and some degradation characterizing historical documents. Some failure cases are presented at the end of the figure where our model fails to retrieve the correct matching word image. This can occur as a result of the stroke corruption of some words like the word 'Waggons' at the fifth row where the last letter 's' is not clearly noticeable. Hence, the model faces an ambiguity in deciding whether this letter is present at the end of the word. In other cases, this can be attributed to the lexical complexity of the word image. For instance, the word "Sergeant" at the sixth row is well retrieved at the top 1 and 2 instances, but it was then conflicted with the word "Regiment" which makes the model not able to decide about some regions of the word image. In other cases, the model is missing the appropriate instances because of the scarcity of this image's representation in the used dataset as is the case of the seventh row of Fig.~\ref{fig:qualitative_result}. 
However, our model was able to retrieve the unique right instance in the second rank for the example shown in the eighth row.
To summarize, we conclude that the ambiguity and the complexity of the script writing and the close similarity among different words are the principal causes of the failures.

\FloatBarrier
\FloatBarrier

\subsection{Ablation study}\label{sec_ablation_study}
 
In preliminary research, we have conducted different experiments to first select the best architecture and the optimal parameters leading to the most efficient \textit{ST-KeyS} model, then to compare the proposed method to different self-supervised/supervised methods.  These investigations are detailed in this section.
 
- \noindent \textbf{Investigating the design choice:} In here, we discuss the self-supervised pretraining and the different downstream task options. We performed several experiments to prove the efficiency of our {ST-KeyS} model that consists of two phases:  a self-supervised pretraining phase followed by a downstream task (supervised fine-tuning phase). The fine tuning stage is further composed of siamese network architecture followed by PHOC embedding (SNN-PHOC). Thus, we study the effectiveness of the pretraining, as well as the utility of the fine-tuning in the chosen order.  The obtained results of this investigation on the three datasets are presented in Table~\ref{tab:resuts_GW_downstreams}, where we studied different scenarios. 
Particularly, we compare the efficiency of the proposed model while using different downstream tasks such as 1) a siamese architecture followed by a PHOC alignement (SNN-PHOC), 2) PHOC followed by a siamese architecture (PHOC-SNN), 3) the use of PHOC representations exclusively and 4) the use of a siamese network uniquely (SNN). Also, we check the utility of the pretraining by comparing it with a supervised-only approach.

As it can be seen from the table, the best performance by our model is with the setting  SNN-PHOC.  We obtain using the GW and AK databases 95.70\% and 85.16\%, respectively. For the BOT dataset, the performance is beyond 59\%.
Hence, we can conclude that the use of a SNN and PHOC downstream task at the fine-tuning stage is the better option, especially in the order SNN-PHOC (first train the SNN and then train with PHOC). The main justification behind this lies in the fact that the siamese network has the capacity to extract relevant deep features leading to a relevant representation of the word image and that is subsequently processed by the PHOC-based encoder allowing the contextual content of the word image to be promoted. 

Moreover, we conducted an evaluation of the self-supervised approaches compared to the supervised ones. As it can be depicted from Table~\ref{tab:resuts_GW_downstreams},  the use of the self-supervised approach followed by a downstream task (SNN) performs better than using a supervised approach based on a siamese architecture (row 5 vs. row 4). A performance of 92.48\% is reached using a pretraining stage while it was at 69.98\% when performing supervised approach on the GW database. Therefore, we can affirm the importance of the pretraining stage in a self-supervised fashion to improve the word spotting results.

Finally, we explored the relevance of the pretraining stage in a self-supervised way without any labeled data. We observe (from the last row in Table~\ref{tab:resuts_GW_downstreams}) that we obtained interesting results from only the image representations. For instance, we achieved a performance of 61.85\% using the GW dataset which is good and even better than some learning-free word spotting methods.
 
\FloatBarrier
\begin{table}[ht]
\caption{{Evaluation of different downstream tasks for word spotting performed on the GW, BOT and AK datasets. (results are presented in term of  mAP\%  metric) }}
\centering
\begin{tabular}{|p{4.5cm}|p{2.3cm}|p{1.2cm}|p{1.2cm}|p{1.2cm}|}
\hline
\textbf{Approach}     & Downstream Task &  GW & BOT & AK \\ \hline \hline
Self-Supervised   &  SNN-PHOC &   \textbf{95.70}  &  59.10           & \textbf{85.16} \\ 
Self-Supervised   &  PHOC-SNN &   92.62           &  \textbf{59.30}  & 82.80  \\ 
Self-Supervised   &  PHOC     &   92.50           &  59.17           & 82.53 \\    
Self-Supervised   &  SNN      &   92.48           &  58.06           & 81.51 \\ \hline
Supervised (SNN)  &  --       &   69.98           &  $NP$              & $NP$     \\ \hline
Self-Supervised (encoder) & -- &  61.85           & 43.79            & 58.63   \\ 
\hline 
\end{tabular}

\textit{NP : Not Performed }
\label{tab:resuts_GW_downstreams}
\end{table}
\FloatBarrier

- \noindent \textbf{Evaluation of the model utility in the low resource scenario:}  This evaluation is based on siamese downstream task using different annotated data amounts. To further prove the importance of the self-supervised approach followed by a downstream task according to the amount of annotated data utilized in the fine-tuning phase, we have conducted a study while varying the amount of labeled data to simulate a low resource scenario. From Table~\ref{tab:resuts_GW_different_amounts_selfsiam}, we remark that we have achieved a good result using merely 20\% of the labeled data which is relatively comparable to the use of 60\% of annotated data and even to the utilization of the entire annotated data provided in the GW database. This confirms the utility of our approach, especially for the low resource datasets, when the size of the labeled data is limited.

\FloatBarrier
\begin{table}[ht]
\caption{{Evaluation results (mAP \%) of the word spotting task performed on the GW dataset for the Self-supervised approach with a siamese downstream task using different data amounts.}} 
\centering
\begin{tabular}{|p{4cm}|p{4cm}|}
\hline
Annotated Fraction\% &      mAP\% \\ \hline
20     & 80.98 \\   
40     & 85.06 \\  
60     & 86.88 \\   
100    & 92.48 \\
\hline 
\end{tabular}
\label{tab:resuts_GW_different_amounts_selfsiam}
\end{table}
\FloatBarrier

- \noindent \textbf{Model's parameters optimisation:} To perform the proposed word spotting method, we have conducted extensive preliminary experiments to set up the suitable parameters for the model configuration that leads to the best performance. Therefore, we start our experiments by selecting the optimum patch size and model architecture.

In our experiments, we define three types of model's variants which are \textit{ST-KeyS-Base-16}, \textit{ST-KeyS-Base-32} and \textit{ST-KeyS-Small-16}, as enlisted in Table~\ref{tab:model_variant}. Obviously, implementing a larger model requires more computational memory and learning time as the number of model parameters increases. Thus, a trade-off between the size of the model and its performance must be considered. 

\FloatBarrier
\begin{table}[ht]
\caption{{Details of our model variants (EL: Encoder Layers, EAH: Encoder Attention Heads, DL: Decoder Layers, DAH: Decoder Attention Heads).}}
\centering
\begin{tabular}{|p{3.2cm}|p{1cm}|p{1cm}|p{1cm}|p{1cm}|p{1.1cm}|p{1.2cm}|}
\hline
Model & EL & EAH & DL & DAH & Patch & Params \\ \hline
\textit{ST-KeyS-Base-16}   & 12 & 6 & 4 & 6 & $16\times16$   &  9.3M           \\ 
\textit{ST-KeyS-Base-32}   & 12 & 6 & 4 & 6 & $32\times32$   &  9.5M           \\  
\textit{ST-KeyS-Small-16}  & 6  & 3 & 4 & 3 & $16\times16$   &  1.2M           \\ 
\hline
\end{tabular}
\label{tab:model_variant}
\end{table}
\FloatBarrier

In Table~\ref{tab:resuts_patch_GW}, we present the results of the Self-supervised approach using only the pretraining representations (without downstream task) from the three defined model variants. We note that the performance of the word spotting method is improved when using a patch size of $16\times16$ instead of using a patch size of $32\times32$ for the same encoder-decoder architecture (row 1 vs. row 2 in Table~\ref{tab:resuts_patch_GW}). The explanation behind this behavior is that, by utilizing a smaller patch size, we  make each patch of the image have more local information during self-attention. Therefore, the model is able to use more and finer information during the word spotting process with a patch size of $16\times16$. 
However, in the case of using a smaller patch size such as $8\times8$, it will lead to the augmentation of the model parameters which results in the requirement of more computation resources. For that reason, we have discarded this setting. Also, we remark that the use of a model having a large number of layers and heads leads to a significant improvement in the word spotting results (row 1 vs. row 3 in Table~\ref{tab:resuts_patch_GW}). Since we attempt to achieve a compromise between these hyper-parameters and resource consumption while maintaining good performance, we selected the \textit{ST-KeyS-Base-16} model's setting as the base model architecture in all our performed experiments.

\FloatBarrier
\begin{table}[ht]
\caption{{Evaluation of the Self-supervised approach (without a downstream task) for word spotting task in terms of mAP\% on the GW database using different model variants.}}
\centering
\begin{tabular}{|p{4cm}|p{1.5cm}|}
\hline
Model    & mAP \% \\ \hline
\textit{ST-KeyS-Base-16}    
& 61.85  \\ 
\textit{ST-KeyS-Base-32}      
&  40.29  \\  
\textit{ST-KeyS-Small-16}   
& 37.53  \\  
\hline
\end{tabular}
\label{tab:resuts_patch_GW}
\end{table}
\FloatBarrier

\section{Conclusion and Future Work}\label{sec_conclusion}

In this paper, we proposed a novel self-supervised approach for keyword spotting called \textit{ST-KeyS}. Our method is based on a pure vision transformer architecture without any CNN layer, and it is composed of a pretraining phase and a fine-tuning phase. In the pretraining stage, a masking-recovering autoencoder is used to learn useful representations from the unlabeled word images. In the fine-tuning stage, a two steps strategy is employed to extract further promoted representations for the spotting task. To produce more robust and meaningful features, a siamese approach is first utilized to embed the images visually, followed by an alignment approach to the  PHOC attributes produced from the text.

The underlined architecture deals with the data scarcity issue since we have restricted our study to the available data provided in the evaluation datasets without using synthetic data or a more extensive dictionary.
Extensive experiments have been made to validate the design choice of our proposal and evaluate it according to the state-of-the-art. The experiment results demonstrated the efficiency of self-supervised learning for word spotting by breaking with the need to have large annotated datasets. We have shown that our model allows achieving close, and sometimes better, performance to fully supervised approaches based on large synthetic datasets despite our restriction to limited data amounts and closed vocabulary.

In the future, we will enhance the performance of our proposed word spotting method by adding an n-gram language model to re-rank the retrieved word image list. We would also like to apply the self-supervised approach in a segmentation-free fashion and for other connected fields, such as the handwriting recognition area.

\section*{Acknowledgement}

This work has been supported by the DocPRESERV (Preserving \& Processing Historical Document Images with Artificial Intelligence) project, funded by the Swedish Foundation for International Cooperation in Research and Higher Education (STINT), grant AF2020-8892.

\bibliography{main.bib}

\end{document}